\documentclass[10pt, a4paper, conference, compsocconf]{IEEEtran}
\ifCLASSINFOpdf
\else
\fi
\usepackage{graphicx}
\usepackage{amsmath}
\usepackage{cite}
\usepackage[colorlinks=true,linkcolor=blue,citecolor=blue]{hyperref}
\begin{document}
%
\title{On color image quality assessment using natural image statistics}


\author{\IEEEauthorblockN{Mounir Omari$^1$, Mohammed El Hassouni$^1$, Hocine Cherifi$^2$ and Abdelkaher Ait Abdelouahad$^1$}
\IEEEauthorblockA{$^1$LRIT URAC 29, University of Mohammed V-Agdal, Rabat, Morocco.\\
$^2$ LE2I UMR 6306 CNRS, University of Burgundy, Dijon, France.\\
mouniro870@gmail.com, mohamed.elhassouni@gmail.com, hocine.cherifi@u-bourgogne.fr, a.abdelkaher@gmail.com}
}


%


\maketitle

\begin{abstract}
Color distortion can introduce a significant damage in visual quality perception, however, most of existing  reduced-reference quality measures are designed for grayscale images. In this paper, we consider a basic extension of well-known image-statistics based quality assessment measures to color images. In order to evaluate the impact of color information on the measures efficiency, two color spaces are investigated: RGB and CIELAB. Results of an extensive evaluation using TID 2013 benchmark demonstrates that significant improvement can be achieved for a great number of distortion type when the CIELAB color representation is used.
\end{abstract}


%
\IEEEpeerreviewmaketitle

\section{Introduction}
\label{intro}
Nowadays, due to the great development of acquisition and reproduction systems, audio visual content is ubiquitous. Unfortunately, processing, coding and transmission algorithms generate various degradations.\\
 Considering these limitations, the assessment of the visual impact of these distortions on images is required. To meet this growing demand, psycho-visual experiments can be performed. Various subjective evaluation processes have been designed in order to quantify the image quality  using  a panel of human observers. However, they are time consuming and difficult to implement. Therefore, objective quality evaluation has been proposed as an alternative solution. Its main goal is to assess the quality of a distorted image using an automatically computed measure, well correlated with human visual judgment. We can identify three main types of objective measures. Full reference (FR) measures which, use the entire original image, as a reference, to estimate the quality of its distorted version. Generally, no information related to the type of distortion is needed. No reference (NR) measures require only the degraded image. However, the type of distortion is generally assumed to be known. They are considered the most attractive from a practical point of view as no side information is needed to grade an image. Reduced reference (RR) measures have been developed to overcome the drawback related to FR and NR measures. This RR measurement can be predicted by only some limited features extracted from both images. Furthermore, no a priori information, about the distortion type, is required. \\
 Recently, a number of authors have successfully introduced RR methods based on: image distortion modeling ~\cite{Ref1,Ref2}, human visual system (HVS) modeling ~\cite{Ref3,Ref4},  and finally natural scene statistics (NSS) modeling ~\cite{Ref5,Ref6}. Our work  fits in  this latter approach. Indeed, natural images statistics are the basic stimuli that our visual system is adapted to. Understanding the way by which statistics change and measuring these changes allows us to predict the visual degradation.  The first measure based on NSS has been introduced by Wang et al~\cite{Ref5}. Known as WNISM, this method is based on the distribution of the steerable pyramid coefficients computed from the images. In order to minimize the side information to be transmitted, a generalized Gaussian distribution model is assumed. Indeed, under this assumption, only the estimated parameters of the distribution are transmitted, instead of the full histogram. At the receiver side the Kullback-Leibler divergence is computed in order to quantify the distortion. This pioneering work has been improved by Li et al~\cite{Ref7}. They introduce a Divisive Normalization Transform which consists on using the same transformation cited above followed by a Gaussianization process of non-Gaussian assumed coefficients distribution. The so-called DNT method has gained in efficiency in terms of correlation with the visual human perceived quality. However, it is much more time consuming, because of the normalization process.\\
 To overcome this drawback, we proposed in a our previous work, a quality measure called EMISM \cite{Ref6}. It uses the Empirical Mode decomposition transform instead of the steerable pyramid. The generalized Gaussian distribution model distribution is assumed and the distortion measure is computed in the same way than for the previous methods, i.e. the Kullback-Leibler divergence. An alternative approach has been proposed by Soundararajan et al~\cite{Ref10}. The so-called RRED method is based on the entropy difference between the wavelet coefficients.\\
All these methods operate on gray level images. While the great majority of images are captured in color, their visual quality is therefore estimated ignoring the color information. According to our knowledge, there is no a real RR measure based on NSS that has been proposed for color image quality assessment.  In an earlier work~\cite{Ref9}, we have attempted to extend the approach based on the marginal distributions to color images. Results were quite promising and motivating enough to study the effect of color on RR measures.
In this paper, we aim to study how color representation can influence the quality assessment process for RR statistical based methods. To meet this objective, first, we choose four RR methods based on NSS in grayscale level, which are WNISM, DNT, RRED and EMISM. Then, we compare their performances in terms of quality assessment to their natural extension to color. We recall that this is a trivial extension by combining the values obtained from the three components by summation. In this study, we consider two influential color spaces, namely RGB and CIELAB. We choose  RGB for its simplicity and suitability to a lot of methods and CIELAB because it is close to human visual perception. To compare these methods, the TID 2013 benchmark is used, due to the abundance of the distortion types it contains. The main contribution of this paper is to give a clear view of the impact of the color information on natural image statistics quality based approaches. Our goal is not to design an optimized quality metric for color images but to pave the way for further improvement.\\
The rest of the paper is organized as follows. Section~\ref{sec:1} presents the selected method and how we extend these methods for color images. Section~\ref{sec:2} concerns the experimental data and the validation protocol. Section~\ref{sec:3} details the experimental results and finally concluding remarks are presented in Section ~\ref{sec:4}.
\begin{figure*}[!t]
\centering
\includegraphics[width=7in]{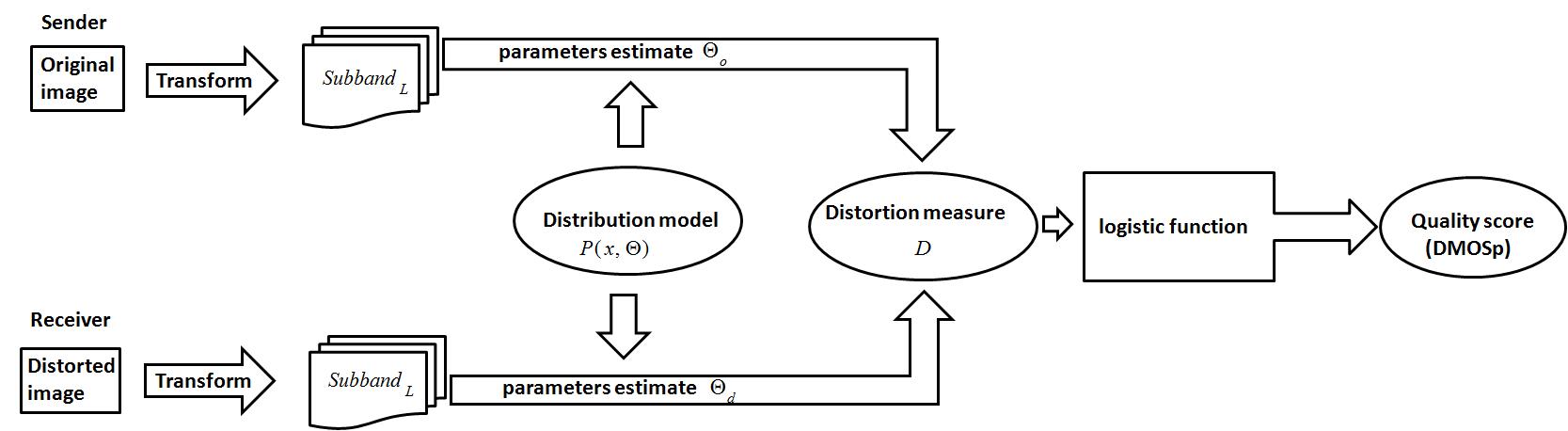}
\caption{The general scheme of natural scene statistics NSS based quality assessment measure.}
\label{fig:1}
\end{figure*}
\section{Background}
\label{sec:1}
Figure~\ref{fig:1} illustrates the general scheme of NSS based quality assessment methods. We distinguish two channels. The sender side is linked to the original image while the receiver side is linked to the distorted image.  On each channel, a transform is performed to get $L$ subbands. Assuming a model $P\left(x|\theta\right)$ for the subbands coefficients distribution, the parameter sets $\theta_{o}$ and $\theta_{d}$ are estimated from the empirical distributions. Note that $\theta_{o}$ and $\theta_{d}$ are respectively the set of parameters vectors $\theta_{o_i}$ and $\theta_{d_i}$ for each subband of the reference image and the degraded image and $i=1,....,L$ is the subband index. At the sender side $\theta_{o}$ is transmitted. At the receiver side a distortion measure $D$ is computed between the original and degraded image subbands distributions. Note that different dissimilarity measures can be used to quantify the differences between the subbands distributions. Finally a non-linear regression using a logistic function is performed in order to compute the predicted MOS. More details are found in the experiments setup section.
\subsection{WNISM}
\label{subsec:11}
WNISM uses the steerable pyramid transform~\cite{Ref19} to get an efficient and accurate linear decomposition of an image into scale and orientation subbands. The generalized Gaussian density (GGD) used for subbands modeling is defined as follows:
\begin{equation}
P(x|\theta)=\frac{\beta}{2\alpha\Gamma\left(\frac{1}{\beta}\right)}\exp\left[-\left(\frac{|x|}{\alpha}\right)^{\beta}\right]
\label{eq:1}
\end{equation}
Where $\alpha$ and $\beta$ represents the scale and the shape parameters respectively that need to be estimated for each subband. The statistical framework to estimate these parameters is the Maximum likelihood, which can be computed efficiently.\\
The overall distortion measure $D$ is based on the sum of the Kullback-Leibler Divergence (KLD) of the $L$ subbands~\cite{Ref5}:
\begin{equation}
D=\log{_{2}}\left(1+\frac{\text{\ensuremath{\sum}}_{i=1}^{L}d_{i}\left(P\left(x|\theta_{o_i}\right)||P\left(x|\theta_{d_i}\right)\right)}{D_{0}}\right)
\label{eq:2}
\end{equation}
Where $D_{0}$ is a constant used to control the scale of the distortion measure and the KLD for the subband $i$ is represented by $d_{i}\left(P\left(x|\theta_{o_i}\right)||P\left(x|\theta_{d_i}\right)\right)$.\\
Note that there exist a closed-form expression for the KLD between two GGD where only the model parameters are involved.
\subsection{DNT}
\label{subsec:13}
Motivated by perceptual and statistical issues, Lee et al~\cite{Ref7} proposed to substitute to the wavelet representation a divisive normalization transform (DNT). It is built upon a wavelet image decomposition, followed by a divisive normalization stage. This process produces approximately Gaussian marginal distributions. Li et al propose an initial model based on the Gaussian Scale Mixture (GSM) for the wavelet coefficients~\cite{Ref12}. It can be expressed as the product of two independent components: $x\dot{=}z.U$, where $\dot{=}$ denotes equality in probability distribution, $U$ is a zero-mean Gaussian random vector with covariance $M$ and $z$ is a scalar random variable called a mixing multiplier. In other words, the GSM model expresses the density of a random vector as a mixture of Gaussians with the same covariance structure $M$ but scaled differently (by $z$). Suppose that the mixing density is $p_{z}(z)$, then the density of $x$ can be written as:
\begin{equation}
P_{x}(x)=\int\frac{1}{[2\pi]^{\frac{N}{2}}\left|z^{2}M\right|^{\frac{1}{2}}}exp-\left(\frac{x^{T}M^{-1}x}{2z^{2}}\right)p_{z}(z)dz
\label{eq:3}
\end{equation}
The DNT coefficients are computed based on the maximum-likelihood estimate of $z$.
After normalization, the DNT marginal statistics can be efficiently modeled by a zero-mean Gaussian distribution. Consequently, the dissimilarity measure between two normalized subband distributions is approximated by the KLD between two Gaussian distributions. In addition, authors propose to add distances based on the difference between the standard deviation, the kurtosis, and the skewness of the DNT coefficients computed from the original and distorted images, respectively. Finally, the overall image distortion measure is represented here as a weighted linear combination of the four distances. This procedure is applied to a single subband, so it is necessary to calculate the sum of all sub-bands to find the overall measure $D$.
\subsection{EMISM}
\label{subsec:12}
In~\cite{Ref11}, we proposed a distortion measure called EMISM that uses the Empirical mode decomposition (EMD) as a transformation domain. The EMD is based on the frequency-time analysis and decomposes each image into a number of intrinsic mode functions (IMFs) and a residue. The basis functions of the EMD derived from the image content allow an adaptive analysis that is more suited to model the visual information. Indeed, in the steerable pyramid representation, the basis functions are fixed and do not necessarily match the varying nature of images. Experimental results have demonstrated that the GGD is a good fit for the IMF distribution. Based on this model, the distortion measure is computed as in WNISM, i.e. the KLD between the IMF statistics.
\subsection{RRED}
\label{subsec:14}
The principal idea underlying RRED as proposed by Soundararajan et al~\cite{Ref10} is to link the distortion to the entropy difference between the wavelet coefficients. The RRED algorithm split the subband in blocks, and then computes the entropy of each block, assuming a GSM model. Then the difference between the entropy of the reference and distorted image is calculated. In this case, $\theta$ is a set of entropies obtained for the blocks. The overall distance $D$ is the summation of the difference of the entropies between the original and degraded blocks. In other case, the distance is the difference between the summation of the entropies of the two chosen subbands.
\subsection{Extension to color image quality assessment}
\label{subsec:25}
As our goal is to investigate the impact of color information rather than optimizing the metrics, we consider a simple adaptation to color images. It is based on the summation of the distortion measures obtained independently from the three-color components. We choose to investigate two color spaces. The RGB color space because it is the most influential additive color space~\cite{Ref14}, and  CIELAB (or L*a*b* or Lab) which is a color-opponent space~\cite{Ref15}. While RGB gives individual values for red, green and blue, L stand for lightness and a and b for the color-opponent dimensions. CIELAB has been designed in order to be a perceptually uniform color space. In other word, a change of the same amount in a color value should produce a change of about the same visual importance.
\begin{table}[h]
\caption{Types of distortion used in TID 2013.}
\label{tab:1}
\vspace{1.5ex}
 \centering %
\begin{tabular}{ll}
\hline\noalign{\smallskip}
 Label & Type of distortion	\\
\noalign{\smallskip}\hline\noalign{\smallskip}
 1& Additive Gaussian noise\\
 2 & Additive noise in color components...\\
 3 & Spatially correlated noise\\
 4& Masked noise\\
 5 & High frequency noise\\
 6 & Impulse noise \\
7&Quantization noise\\
8& Gaussian blur\\
9&Image denoising\\
10&JPEG compression\\
11&JPEG2000 compression\\
12&JPEG transmission errors\\
13&JPEG2000 transmission errors\\
14& Non eccentricity pattern noise\\
15 & Local block-wise distortion of different intensity\\
16&Mean shift\\
17&Contrast change\\
18&Change of color saturation\\
19&Multiplicative Gaussian noise\\
20&Comfort noise\\
21&Lossy compression of noisy images\\
22&Image color quantization with dither\\
23&Chromatic aberrations\\
24&Sparse sampling and reconstruction\\
\noalign{\smallskip}\hline
\end{tabular}
\vspace{1.5ex}
\end{table}
\section{Experimental setup}
\label{sec:2}
\subsection{Dataset}
\label{sec:31}
\begin{figure}[h]
\centering
\includegraphics[width=3in]{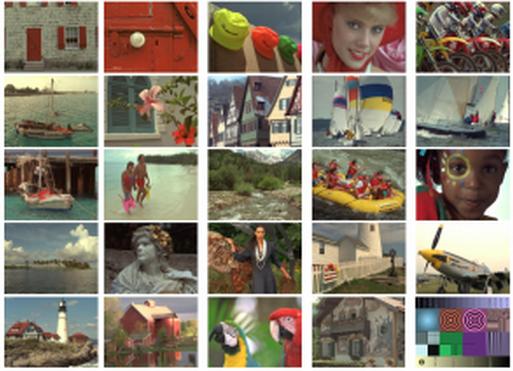}
\caption{The reference images of the TID 2013 dataset.}
\label{fig:2}
\end{figure}

 To test the performances of the measures under investigation, we use the TID 2013 dataset~\cite{Ref21}. This recently released benchmark has been specially designed for quality metrics evaluation. TID 2013 is an improved version of TID 2008 database. It contains 25 reference images shown in figure~\ref{fig:2}, and 3000 distorted images (25 reference images with 24 types of distortions and 5 levels of distortions). These distortions type are listed in table~\ref{tab:1}. The quality of each image in TID 2013 has been graded by the Mean Opinion Score (MOS) and experiments was carried out by 971 observers.

\subsection{Validation protocol}
\label{sec:32}
To compare the proposed measure with the subjective quality score (MOS), we perform a nonlinear regression using a logistic function in order to map the objective and subjective scores~\cite{Ref16}. The logistic function proposed by the Video Quality Expert Group (VQEG) Phase I FR-TV with five parameters is used. The expression of the quality score which is the predicted MOS is given by:
\begin{equation}
\label{prediction}
DMOS_{p}=\beta_{1}\mathrm{logistic}\left(\beta_{2},D-\beta_{3}\right)+\beta_{4}D+\beta_{5}.
\end{equation}
\emph{fminsearch} function is used to estimate the vector $\left(\beta_{1},\beta_{2},\beta_{3},\beta_{4},\beta_{5}\right)$ in the optimization Toolbox of Matlab, and the logistic function is expressed by:
\begin{equation}
\label{logi}
\mathrm{logistic}\left(\tau,D\right)=\frac{1}{2}-\frac{1}{1+\exp\left(\tau D\right)}.
\end{equation}
Where $D$ is the overall measure in equations \ref{prediction} and \ref{logi}.\\
The prediction accuracy together with the prediction monotonicity are computed in order to evaluate the relevance of a quality metric. The prediction accuracy is measured by the Pearson’s linear correlation coefficient (PLCC) as defined in eq. \ref{PLCC}, while the prediction monotonicity is quantified by the Spearman rank-order coefficient SRCC as shown in eq. \ref{SRCC}. Where $i$ denotes the index of the image sample and $N$ is the number of samples.
\begin{figure*}[!ht]
\begin{equation}
PLCC=\frac {\sum_{i=1}^{n}(DMOS(i)-\bar{DMOS})(DMOS_p(i)-\bar{DMOS_p})}{\sqrt {\sum_{i=1}^{n}(DMOS(i)-\bar{DMOS})^{2}} \sqrt{\sum_{i=1}^{n}(DMOS_p(i)-\bar{DMOS_p})^{2}}}\label{PLCC}
\end{equation}
\begin{equation}
SRCC=1-\frac{6\sum_{i=1}^{^{N}}(rank(DMOS(i))-rank(DMOS_p(i)))^{2}}{N(N^{2}-1)}\label{SRCC}
\end{equation}
\end{figure*}
\begin{table*}[!ht]
\caption{Comparison between color and the grayscale of the four methods using the PLCC values and TID 2013 dataset.}
\label{tab:2}
\vspace{1.5ex}
 \centering %
\begin{tabular}{l|lll|lll|lll|lll}
\hline\noalign{\smallskip}
 Label& \multicolumn{3}{c}{WNISM} & \multicolumn{3}{c}{RRED}& \multicolumn{3}{c}{EMISM}& \multicolumn{3}{c}{DNT}\\
 & {Grayscale} & {RGB} & {LAB} & {Grayscale} & {RGB} & {LAB} & {Grayscale} & {RGB} & {LAB}& {Grayscale} & {RGB} & {LAB} \\
\noalign{\smallskip}\hline\noalign{\smallskip}
{1} &0.69&0.67&0.86&0.79&0.80&{0.86}&0.65&0.64&{0.71}&0.48&0.47&{0.78}\\
{2} &0.62&0.53&0.80&0.78&0.76&{0.85}&0.47&0.53&{0.63}&0.26&0.46&{0.67}\\
{3} &0.69&0.69&0.86&0.79&{0.86}&{0.86}&0.61&{0.68}&{0.68}&0.59&0.46&{0.69}\\
{4} &0.68&0.69&{0.73}&0.83&{0.85}&0.59&0.38&0.60&{0.62}&0.19&0.44&{0.59}\\
{5} &0.77&0.77&{0.92}&0.85&0.92&{0.94}&0.78&0.78&{0.79}&0.40&0.57&{0.80}\\
{6} &0.66&0.69&{0.81}&0.69&0.73&{0.86}&0.49&0.46&{0.54}&0.36&0.37&{0.52}\\
{7} &0.65&{0.71}&0.70&0.76&{0.82}&0.76&0.73&{0.83}&0.79&0.27&{0.69}&0.68\\
{8} &0.91&{0.92}&0.90&{0.95}&0.94&0.94&0.88&0.88&{0.89}&0.81&0.72&{0.82}\\
{9} &0.88&{0.89}&0.83&0.94&{0.95}&0.91&0.83&{0.86}&0.81&0.86&{0.88}&0.86\\
{10} &{0.87}&0.86&0.81&0.94&{0.96}&0.89&0.77&{0.82}&0.70&0.53&0.57&{0.77}\\
{11} &0.93&{0.94}&0.89&{0.97}&0.96&0.90&{0.91}&0.90&0.89&{0.91}&0.85&0.90\\
{12} &0.88&{0.90}&0.68&0.89&{0.93}&0.80&0.47&{0.55}&0.46&0.34&0.48&{0.49}\\
{13} &{0.78}&0.77&0.74&0.72&{0.78}&0.70&{0.64}&0.62&0.60&{0.74}&0.54&0.58\\
{14} &0.44&{0.50}&0.46&0.76&0.77&{0.81}&0.36&0.46&{0.56}&{0.60}&0.31&0.40\\
{15} &0.22&0.16&{0.25}&{0.53}&0.51&0.23&0.28&0.20&{0.29}&0.13&{0.46}&0.34\\
{16} &0.48&0.58&{0.73}&0.66&0.71&{0.77}&0.50&0.53&{0.58}&0.31&0.42&{0.65}\\
{17} &{0.68}&0.65&0.38&{0.57}&0.56&0.06&0.68&{0.69}&0.45&{0.40}&0.20&0.39\\
{18} &0.21&0.51&{0.67}&0.06&{0.75}&0.65&0.18&{0.25}&{0.25}&0.26&0.32&{0.48}\\
{19} &0.62&0.59&{0.85}&0.75&0.75&{0.85}&0.58&0.61&{0.68}&0.52&0.44&{0.63}\\
{20} &{0.74}&0.73&0.56&{0.91}&0.90&0.83&0.16&{0.47}&0.45&0.12&0.26&{0.30}\\
{21} &0.73&0.67&{0.91}&0.90&0.93&{0.94}&0.64&0.66&{0.67}&0.32&0.46&{0.70}\\
{22} &0.49&0.48&{0.73}&0.81&{0.86}&0.84&0.77&0.74&{0.78}&0.36&{0.68}&0.57\\
{23} &{0.96}&0.94&0.80&0.95&{0.98}&0.90&0.92&0.92&{0.94}&{0.91}&0.87&0.75\\
{24} &0.92&{0.93}&0.89&{0.97}&{0.97}&0.87&0.91&{0.92}&0.91&0.87&0.82&{0.90}\\
{All} &0.69&0.7&0.74&0.78&0.83&0.78&0.61&0.65&0.65&0.48&0.53&0.64\\
\\
\noalign{\smallskip}\hline
\end{tabular}
\vspace{1.5ex}
\end{table*}
\section{Experimental Results}
\label{sec:3}
The major purpose of our validation work is to verify that using color image representation is beneficiary for assessing quality by comparing the performance of the proposed extension of the selected methods with their grayscale level implementations.
 The values obtained by the PLCC and SRCC on the TID 2013 database are shown in Table~\ref{tab:2} and~\ref{tab:3}, respectively.
\subsection{Color spaces evaluation}
\label{subsec:31}
In this section, we compare the two selected color spaces against the grayscale level. To do so, we use a performance evaluation improvement for each distortion type using :
\begin{equation}
Per= \left(\left(CC_{Grayscale}-CC_{Color}\right)/CC_{Grayscale}\right)*100
\end{equation}
Where  $CC_{Color}$  is a correlation coefficient which can take  PLCC or SRCC values for color based methods, and $CC_{Grayscale}$ can take  PLCC or SRCC values for grayscale level methods. For our comparisons, we tune three classes of improvement. Class 1 (negligible) for which less than $5\%$  is considered, Class 2 (medium) for percentages centered between $5\%$ and $10\%$ and Class 3 (high) for which the percentage is higher than $10\%$. \\
Let's begin with Table~\ref{tab:2} which gives the PLCC values. In order to see the influence of color on the four selected methods, we classify the twenty four distortions types of TID 2013 into two categories. The first concerns types that affect the color (2, 7, 10, 18, 22, 23) and the second covers the rest of distortions.\\
When considering color related artifacts, we begin with the Additive noise in color components. Since it is  modeled in the YCbCr color space, it  has been added to test the quality metric appropriateness to the HVS properties. This allows to not equally perceive distortions in luminance and chrominance components. RGB shows its effectiveness for EMISM and DNT with an improvement higher than $10\%$. However, it fails for  WNISM and RRED. So, the use of steerable pyramid transform and entropy deteriorates the results of RGB. The efficiency of the LAB color space is clearly seen, with  an improvement of class 3 for three methods (WNISM, EMISM and DNT) and  class 2 for (RRED).\\
Quantization noise is a distortion type that has not received too much attention in image visual quality evaluation. Although, this distortion is quite often met in practice and allows to estimate quality of the metrics adequacy with respect to several peculiarities of HVS as color, local contrast and spatial frequency. For this distortion type, we can find an improvement of class 3 for RGB  with (EMISM and DNT), and for LAB  with DNT. While the remaining results of this  type are found in the class 2, except for RRED with LAB which the PLCC value ​​is close to 0.76.\\
Let's turn to JPEG compression. This distortion type is based on lossy compression. Two tasks of degradation have been included, color and  spatial frequency sensitivity. For the common method WNISM, the values become small for both RGB and LAB due to the use of steerable pyramid transform. This degradation affects also RRED and EMISM when the LAB  is used. With less than $5\%$ of improvement, the RRED gives an advantage for RGB. This later   outperforms grayscale for EMISM and RRED with an achievement of class 2. Only for DNT with LAB, the improvement  belongs to the class 3. This small advantage  is caused by the fact that the artifact nature itself reaches the spatial frequency sensitivity.\\
Without a doubt, the improvement for the change of color saturation is needed, hopefully, the RGB and LAB upgrade the graysclae performances with a high percentage of class 3.\\
The distortion type Image color quantization with dither is understood from its name. It converts RGB image to indexed image using dither. The enhancement here is clearly seen for DNT with more than $10\%$, such as WNISM for LAB. As well for RRED and EMISM, the LAB improves grayscale but, with an improvement of class 1. In class 2, only for RRED with RGB, the PLCC value is increased.\\
The last distortion type of color is the chromatic aberrations. It was modeled by a slight mutual shifting of R, G, and B components with respect to each other with further blurring of shifted components. This artifact is successful only for two cases (RRED with RGB, and EMISM with LAB) with less than $5\%$ of enhancement. Despite the effect of this distortion type, the RGB and LAB do not improve the quality score, which lead us to use others color spaces for this distortion type.\\
Focusing now on artifacts for which color have not handled. There is no deterioration for the High frequency noise, which allows analyzing metrics adequateness with respect to local contrast sensitivity and spatial frequency sensitivity of HVS. Meanwhile, the values of WNISM with RGB and grayscale are equal. On the other hand, RGB and LAB slightly increase the quality score for EMISM. The improvement for DNT,  RRED and WNISM with LAB belongs to class 3, while the RGB promotes the quality score of RRED with an enhancement of classe 2.\\
The Additive Gaussian noise is commonly modeled as a white Gaussian noise. For this artifact, the LAB boosts the gray-scale for WNISM and DNT with a percentage of class 3. For the class2, the remaining values of LAB are located. When the RGB color space is recalled, the improvement reaches precisely a percentage of $0.88\%$ with RRED, but, it fails for the rest, which is a benefit for LAB for this distortion type.\\
The next distortion type is the spatially correlated noise. From its name, it allows analyzing metrics adequateness with respect
to local contrast sensitivity and spatial frequency sensitivity of HVS. Here, the enhancement is throughout, expect for DNT with RGB, which requires the use of color spaces for this type.\\
For five distortion types (Masked noise, Impulse noise, Non eccentricity pattern noise, Mean shift, Lossy compression of noisy images), the integration of color is needed. The domination of RGB and LAB on gray-scale is clearly seen for Mean shift, and only one individual failure for the color spaces is detected for the remaining methods.\\
For the Gaussian blur, which is an important type of distortion often met in practical applications and frequently  included in studies dealing with visual quality metrics. The enhancement for this artifact is accomplished just for three cases (WNISM with RGB, EMISM and DNT with LAB) with an improvement of class 1. For the rest, the deterioration is achieved everywhere.  This result shows that the addition of color does not always enhances the gray-scale.\\
In further debate, the image denoising is a residual distortion resulted after applying different denoising procedures (filters). It is based on spatial frequency and local contrast. It showed, as like as the Gaussian blur that the use of color does not always improves the result, with only a percentage of class 1. The achievement noticed was with RGB for all methods and with LAB for DNT. This result gives a slightly advantage to RGB compared to LAB for this artifact.\\
The JPEG transmission errors is a distortion type based on data transmission to give an eccentricity. The values of DNT and EMISM with RGB showed the high efficiency of color spaces for this artifact with a percentage of class 3. We get an improvement for the RGB for WNISM and RRED with a percentage of class 1. However,  LAB fails for the remaining methods. These results gives a major advantage to RGB against LAB for this artifact.\\
As the bill goes forward, we will be looking closely at Local block-wise distortions of different  intensity. This artifact assumes that in case of compact impulse-like distortions, HVS does not react to distortion on single pixel but mainly to an area (percentage of pixels). Enhancing the gray-scale accuracy with LAB is clearly seen for three methods (WNISM and DNT in class3, and EMISM in class1). In the case of RGB, it gives an enhancement only for DNT, which proves the superiority of LAB against RGB for this distortion type.\\
In the following, the next distortion type to discuss is the  Multiplicative Gaussian noise. It has been chosen to represent a wide class of distortions caused by signal-dependent noise which takes place in many modern applications of CCD sensors, in medical, ultrasound and radar imaging. The results of LAB proved the effectiveness of this color space for this artifact with more than $10\%$  for all four methods. With an improvement  of class 3 that RGB increases the quality for EMISM. The quality scores show also  the  inefficiency of RGB color space for steerable pyramid transform and DNT coefficients.\\
The comfort noise has been added into consideration to take into account a specific feature of human vision that it practically does not matter for it what realization of the noise takes place for a given image. The color information requirement is clear for EMISM and DNT to increase the accuracy of this artifact, hopefully, we have this improvement with a percentage of class 3. But for the remaining methods, we have an important deterioration caused by the entropy and steerable pyramid facts.\\
We focus now on the Sparse sampling and reconstruction. The method of compressive sensing image reconstruction has been used by this artifact in generating distorted images. The improvement with LAB affected only DNT with $3.21\%$. Contrariwise for the RGB, it improves the results of EMISM and WNISM with less than $1\%$. The values of RRED are equal, and for DNT, RGB deteriorates the gray-scale result.\\
Finally, for the remaining distortion types. The integration of color degrades the quality scores, which lead us to think about this fact.\\
To summarize, the improvement of the four methods for Impulse noise, Mean shift, and  Change of color saturation is clearly good with a percentage higher than $10\%$ for the LAB color space. Due to the nature of these artifacts, they affect the colors part directly, especially for the Change of color saturation, which focused on the change in the saturation of the color components. To strengthen this, the RGB gives an obvious improvement to this type of distortion which belongs to class 3 for all four methods. Unfortunately, for the contrast change, we have a deterioration for all four methods.\\
To sum up, the RGB improves the quality for 22 cases, and the LAB upgrades for 41 cases with an achievement of class 3. For improvement of class 2, the RGB and LAB enhance gray-scale respectively for 12 cases, and 9 cases. The class1 contains 28 achievement for RGB, and 12 enhancement for LAB. This result shows that the LAB and the RGB are effective for 62 cases, meanwhile, the competition between these color spaces gives an major advantage to LAB of class 3 with more than $10\%$.  However, for class 1 and class 2, the RGB outperforms LAB. This  proves the effectiveness of the use of color information to these four methods. Unfortunately, these two color spaces have failed for 34 cases for each one of them.\\
\begin{table*}[!t]
\caption{Comparison between color and the grayscale of the four methods using the SRCC values and TID 2013 dataset.}
\label{tab:3}
\vspace{1.5ex}
 \centering %
\begin{tabular}{l|lll|lll|lll|lll}
\hline\noalign{\smallskip}
 Label& \multicolumn{3}{c}{WNISM} & \multicolumn{3}{c}{RRED}& \multicolumn{3}{c}{EMISM}& \multicolumn{3}{c}{DNT}\\
 & {Grayscale} & {RGB} & {LAB} & {Grayscale} & {RGB} & {LAB} & {Grayscale} & {RGB} & {LAB}& {Grayscale} & {RGB} & {LAB} \\
\noalign{\smallskip}\hline\noalign{\smallskip}
{1}&0.69&0.69&{0.87}&0.85&0.85&{0.86}&0.66&0.66&{0.73}&0.66&0.69&{0.78}\\
{2}&0.62&0.56&{0.82}&0.79&0.76&{0.84}&0.53&0.56&{0.61}&0.6&0.59&{0.65}\\
{3}&0.68&0.69&{0.86}&0.85&0.85&{0.86}&0.61&0.67&{0.69}&0.64&{0.69}&{0.69}\\
{4}&0.65&0.66&{0.75}&{0.81}&{0.81}&0.68&0.37&0.59&{0.63}&0.54&{0.64}&0.61\\
{5}& 0.75&0.74&{0.9}&0.89&0.88&{0.9}&0.75&0.75&{0.79}&0.75&{0.79}&{0.79}\\
{6}&0.68&0.71&{0.82}&0.79&0.8&{0.86}&0.49&0.48&{0.59}&0.52&{0.58}&0.55\\
{7}&0.64&{0.72}&0.7&{0.83}&0.81&0.76&0.71&{0.79}&0.75&0.68&{0.69}&0.65\\
{8}&{0.92}&{0.92}&0.9&{0.97}&0.96&0.94&0.88&{0.89}&0.88&0.83&0.84&{0.88}\\
{9}&{0.84}&{0.84}&0.83&{0.92}&{0.92}&0.91&0.78&{0.8}&0.79&0.83&{0.85}&0.82\\
{10}&{0.84}&0.83&0.79&{0.93}&0.91&0.87&0.71&{0.76}&0.74&{0.77}&0.75&0.74\\
{11}&{0.91}&{0.91}&0.87&{0.95}&{0.95}&0.88&{0.86}&{0.86}&0.84&0.91&{0.92}&0.91\\
{12}&0.81&{0.83}&0.64&0.85&{0.9}&0.77&0.43&{0.54}&0.48&{0.72}&0.62&0.55\\
{13}&{0.76}&0.75&{0.76}&{0.79}&0.78&0.7&{0.65}&0.62&0.63&0.74&{0.75}&0.55\\
{14}&0.45&0.47&{0.54}&0.78&0.78&{0.82}&0.34&0.45&{0.58}&0.61&0.66&{0.68}\\
{15}&0.25&0.15&{0.29}&{0.55}&0.52&0.19&0.01&0.2&{0.31}&{0.35}&0.1&0.19\\
{16}&0.45&0.52&{0.74}&0.61&0.67&{0.74}&0.48&0.51&{0.59}&0.48&0.57&{0.73}\\
{17}&0.52&{0.53}&0.33&{0.34}&0.32&0.03&0.54&{0.55}&0.54&{0.88}&0.13&0.29\\
{18}&0.16&0.54&{0.63}&0.05&{0.75}&0.72&0.12&0.33&{0.37}&0.57&0.56&{0.71}\\
{19}&0.6&0.59&{0.85}&0.79&0.79&{0.84}&0.58&0.6&{0.69}&0.55&0.62&{0.64}\\
{20}&{0.75}&0.74&0.61&{0.9}&0.89&0.85&0.16&0.46&{0.48}&0.58&0.51&{0.66}\\
{21}&0.72&0.69&{0.91}&{0.92}&{0.92}&{0.92}&0.63&0.67&{0.7}&0.7&{0.74}&0.73\\
{22}&0.48&0.46&{0.73}&{0.86}&0.84&0.84&0.75&0.7&{0.78}&0.72&{0.73}&0.53\\
{23}&{0.85}&0.71&0.82&0.89&{0.9}&0.87&0.74&0.77&{0.79}&0.84&0.66&{0.68}\\
{24}&0.91&{0.92}&0.88&{0.95}&{0.95}&0.87&0.9&0.91&{0.92}&0.87&0.87&{0.9}\\
{All}&0.66&0.67&0.74&0.79&0.81&0.77&0.57&0.63&0.66&0.68&0.65&0.66\\
\noalign{\smallskip}\hline
\end{tabular}
\vspace{1.5ex}
\end{table*}
Now, we look at monotonicity as a comparison criterion of the grayscale level against the RGB and LAB color spaces. Table~\ref{tab:3} reports the results obtained for TID 2013 dataset. \\
In the case of color artifact, the improvement for the SRCC is less than for the PLCC.\\
For example,  for RGB color space, DNT has failed for Additive noise in color components, JPEG compression, Change of color saturation, and Chromatic aberrations, with a percentage of class1. The enhancement has affected only two distortion types (Quantization noise, Image color quantization with dither). For LAB, when the Quantization noise, JPEG compression, Image color quantization with dither, and Chromatic aberrations are considered, the quality scores of grayscale  have deteriorated. These results give an advantage to PLCC for the DNT when the color information is involved.\\
In view of the EMISM, the achievement is all over the color artifact for both RGB and LAB, except for Image color quantization with dither when the RGB is recalled, but here we have more values ​​of class 1 and class 2 than class 3.\\
Considering RRED, the enhancement is clearly needed for Change of color saturation, hopefully, we have here a great improvement of class 3 with a SRCC higher than 0.7. For the remaining results, only in two cases (Additive noise in color components with LAB and Chromatic aberrations with RGB) that the color information improves the quality scores with percentages of class 2 and class 1 respectively.\\
The RGB enhances just the Quantization noise and the Change of color saturation with more than $10\%$. The LAB is performing better for four cases (class3 (Additive noise in color components, Change of color saturation, Image color quantization with dither), class2 (Quantization noise)).\\
The results of color artifacts show that the improvement is dependent on the nature of the correlation coefficient, the enhancement of PLCC is more efficient than the SRCC for the four methods.\\
Let's consider the others types of artifacts. When the DNT is accomplished, we find an improvement of class3, for RGB with four artifacts (Masked noise, Impulse noise, Mean shift, and Multiplicative Gaussian noise), and for LAB with five artifacts (Masked noise, Non eccentricity pattern noise, Mean shift, Multiplicative Gaussian noise, and Comfort noise). Otherwise, the RGB improves gray-scale for four types of distortion (Spatially correlated noise, High frequency noise, Non eccentricity pattern noise, and  Lossy compression of noisy images) with a percentage of class 2, likewise for LAB for the following artifacts (Spatially correlated noise, High frequency noise, Quantization noise, and Gaussian blur). For class1 achievement percentage, the RGB gives an improvement for five artifacts (Additive Gaussian noise, Gaussian blur, Image denoising, JPEG2000 compression, JPEG2000 transmission errors), while LAB outperforms gray-scale for two cases (Lossy compression of noisy images, Sparse sampling and reconstruction).\\
In case of EMISM, the RGB has failed for two artifacts (Quantization noise, JPEG2000 transmission errors), such as LAB for (JPEG2000 compression, JPEG2000 transmission errors). For this method, the color spaces are more efficient for SRCC than PLCC.\\
The results of the remaining methods show that the improvement required in order to reach the goal of class3 has affected PLCC more than SRCC.\\
Experiments performed with PLCC on standard color images for wide variety of distortion in TID 2013 database indicate that the effect of color in image quality metric is outperforming the grayscale level is such types of distortion that affect color component. In other hand, in some distortion types, we do not need the information of color, contrariwise, the later degrades the performances. It is concluded that  RGB and LAB color spaces are showing very good correlation against gray-scale in some cases. However, the correlation values are very low across various distortions as well as showing the consistent results at different levels of distortions.
\subsection{Comparison between the four methods}
\label{subsec:32}
The correlation between the methods in terms of color spaces and types of distortion is a very important issue. Table \ref{tab:4} and \ref{tab:5} report respectively the correlation between results shown in tables \ref{tab:2} and \ref{tab:3}.\\
 Table  \ref{tab:4} shows that the correlation between the four methods in LAB color space is impressive except for EMISM and RRED methods which is equal to 0.59. For the RGB, the values are swinging between 0.7 and 0.8 which is good as a correlation index, expect between EMISM and RRED that the value is close to 0.62. Globally, the results of grayscale clearly demonstrates the superiority of the correlation between DNT and RRED on LAB and RGB color spaces.\\
In table \ref{tab:5}, we can see that the correlations  between results obtained  by WNISM and RRED for color and grayscale are almost the same.  Comparing DNT and EMISM, we find that LAB is the suitable color space for these methods with a correlation rate of 0.85. The RGB color space records an increase of more than 0.05 for WNISM and EMISM methods when comparing with grayscale and LAB. The grayscale fails for the DNT and RRED with a 0.46 as a correlation. The LAB dominates with a rate of  0.83 for DNT and WNISM methods.\\
To summarize,  this correlation study on the obtained results  shows that in general the LAB color space is more suitable for the extension of the presented methods.
\begin{table}[h]
\caption{Correlations between results obtained in table II.}
\label{tab:4}
\vspace{1.5ex}
 \centering %
\begin{tabular}{l|l|l|l}
\hline\noalign{\smallskip}
 & {Grayscale} & {RGB} & {LAB}\\
\noalign{\smallskip}\hline\noalign{\smallskip}
{(WNISM,RRED)} &0.82&0.72&0.86\\
{(DNT,EMISM) } &0.78&0.78&0.7\\
{(WNISM,EMISM)} &0.74&0.77&0.75\\
{(DNT,RRED)} &0.36&0.89&0.82\\
{(EMISM,RRED)} &0.62&0.62&0.59\\
{(DNT,WNISM} &0.69&0.75&0.81\\
\noalign{\smallskip}\hline
\end{tabular}
\vspace{1.5ex}
\end{table}
\begin{table}[h]
\caption{Correlations between results obtained in table III.}
\label{tab:5}
\vspace{1.5ex}
 \centering %
\begin{tabular}{l|l|l|l}
\hline\noalign{\smallskip}
 & {Grayscale} & {RGB} & {LAB}\\
\noalign{\smallskip}\hline\noalign{\smallskip}
{(WNISM,RRED)} &0.82&0.8&0.81\\
{(DNT,EMISM) } &0.75&0.76&0.85\\
{(WNISM,EMISM)} &0.71&0.78&0.71\\
{(DNT,RRED)} &0.46&0.63&0.61\\
{(EMISM,RRED)} &0.6&0.69&0.62\\
{(DNT,WNISM)} &0.64&0.56&0.83\\
\noalign{\smallskip}\hline
\end{tabular}
\vspace{1.5ex}
\end{table}
\section{Conclusion}
\label{sec:4}
The intention of this work was to study the impact of color in the process of image quality assessment. Four approaches are taken into consideration by extension to color which are WNISM, EMISM, DNT and the RRED.\\
From the presented results, we saw clearly that color and especially LAB space can improve and facilitate the process of the quality assessment and judgment.\\
As a future work, we tend to extend to other color spaces and also see what is the color component (luminance or chrominance) that most influences the judgment of quality.  Furthermore, we plan to investigate alternative models for the marginal distribution of wavelet and BEMD coefficients to introduce the dependencies between color components. In addition, as this study revealed poor performances, we plan to pay more attention to improve the correlation coefficients between the objective and subjective scores.
\section*{Acknowledgment}
This work has been supported by the project CNRS-CNRST STIC 02/2014.





%

\end{document}